\documentclass[review]{elsarticle}

\usepackage{hyperref}

\usepackage{xcolor}
\usepackage{textcomp}
\usepackage{amsmath,amssymb,amsfonts}

\journal{Journal of \LaTeX\ Templates}

\begin{document}

\begin{frontmatter}

\title{Entropy Guided Adversarial Model for Weakly Supervised Object Localization}
\tnotetext[mytitlenote]{Fully documented templates are available in the elsarticle package on \href{http://www.ctan.org/tex-archive/macros/latex/contrib/elsarticle}{CTAN}.}

\author[mymainaddress]{Sabrina Narimene Benassou}
\ead{benassou.narimene@hit.edu.cn}

\author[mysecondaryaddress]{Wuzhen Shi}
\ead{wzhshi@szu.edu.cn}

\author[mymainaddress]{Feng Jiang\corref{mycorrespondingauthor}}


\cortext[mycorrespondingauthor]{Corresponding author}
\ead{fjiang@hit.edu.cn}

\address[mymainaddress]{92 Xidazhi Street, Nangang District, Harbin City, Heilongjiang Province, Harbin Institute of Technology}
\address[mysecondaryaddress]{College of Electronics and Information Engineering, Shenzhen University No. 3688, Nanhai Avenue, Nanshan District, Shenzhen, China}

\begin{abstract}
Weakly Supervised Object Localization is challenging because of the lack of bounding box annotations. Previous works tend to generate a class activation map i.e CAM to localize the object. Unfortunately, the network activates only the features that discriminate the object and does not activate the whole object. Some methods tend to remove some parts of the object to force the CNN to detect other features, whereas, others change the network structure to generate multiple CAMs from different levels of the model. In this present article, we propose to take advantage of the generalization ability of the network and train the model using clean examples and adversarial examples to localize the whole object. Adversarial examples are typically used to train robust models and are images where a perturbation is added. To get a good classification accuracy, the CNN trained with adversarial examples is forced to detect more features that discriminate the object. We futher propose to apply the shannon entropy on the CAMs generated by the network to guide it during training. Our method does not erase any part of the image neither does it change the network architecure and extensive experiments show that our Entropy Guided Adversarial model (EGA model) improved performance on state of the arts benchmarks for both localization and classification accuracy.
\end{abstract}


\end{frontmatter}


\section{Introduction}
\label{sec:introduction}
Weakly supervised learning has gained a lot of popularity during these past few years, especially since Zhou et al \cite{b2} propose the use of Class Activation Map (CAM) to localize objects without bounding boxes annotations. Since that, CAM was extensively used for object localization \cite{b1,b3,b4,b5,b11,b12,b19}, object detection \cite{b26,b27,b28,b29,b30,b31}, image segmentation \cite{b32,b33,b34}, etc. However, not the whole object is highlighted on the CAM, because the network learns the features that discriminate the most the object. Some works have been proposed to deal with this probelm, we can classify them into two approaches. The first approach \cite{b1,b3,b4,b19} consists of hiding a part or some parts of the image during training to force the CNN to detect the full object, nevertheless this approach has a drawback, when the most discriminative part is removed the CNN tends to learn regions of the image that does not belong to the object (as water or tree branches in CUB dataset) because they appear frequently in the training samples \cite{b4,b23}. Furthermore, removing some parts of the image results in information loss for the network and hence decreases its recognition ability \cite{b19}. The second approach \cite{b5,b11,b12} consists of activating different parts of the object by generating multiple CAMs from different levels of the network. These methods however, require a modification in the network structure by plugging some layers or some blocks to the network, which is not always intuitive to design or to generalize for different network structures. In this paper, we propose to take advantage of the generalization ability of the network and propose to use Adversarial Learning (AL) \cite{b13,b16} and entropy \cite{b22,b21} to tackle the limitations of the two approches. Our method does not modify the network backbone, which make the implementation easier, and because the whole images are used for training, there is no information loss for the model.

\begin{figure}[t]
  \centerline{\includegraphics[width=0.5\textwidth]{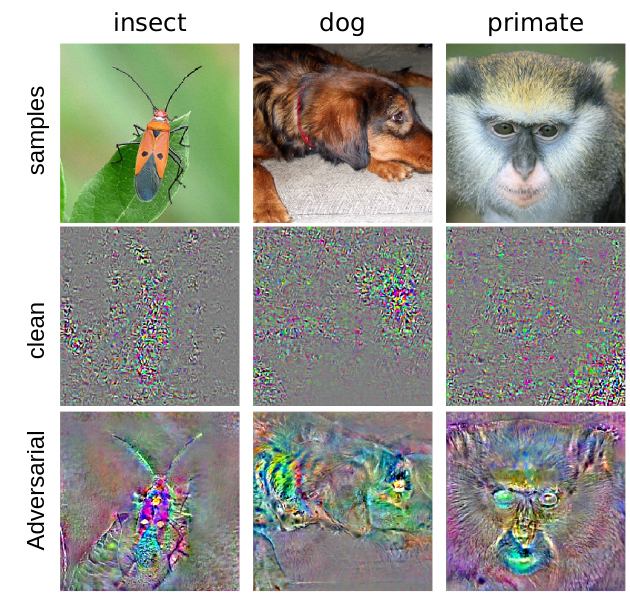}}
\caption{The difference between the features that affect the classification prediction when a model is trained with clean examples and a model trained with adversarial (noisy) examples \cite{b14}}
\label{ref:dif}

\end{figure}

Adversarial learning was originally made to prevent adversarial attacks by training the model with adversarial examples that are generated by introducing some perturbations to clean images. In \cite{b14}, Tsipras et al. visualize the gradient of the loss on ImageNet dataset and have shown the features that affect the classification prediction of a CNN trained with clean examples and a CNN trained with adversarial (noisy) examples. As we can see in Fig. \ref{ref:dif}, the CNN trained adversarially, detect more relevant and clearer features such as edges and borders (line 3) than the standard one, that detects noisy features (line 2). This leads us to the following conclusion; to get a good classification accuracy, the CNN trained only with clean images activate the features that better discriminate the object e.g the head of a bird. But if we futher train the CNN with images where we add a small perturbation, the network is forced to look for more relevant features to better recognize the object e.g the body of the bird. By training a model with both clean and adversarial examples, the generated CAM will activate more discriminative pattern to recognize the object without erasing any part of the image or changing the network architecture. Furthermore, Goodfellow et al. \cite{b16} advice the use of a mixture of clean and adversarial examples to train the CNN. Data augmentation such as rotation or translation will not occur naturally and treating adversarial examples as data augmentation will in fact regularize the network. 

As adversarial learning detects more discrimative features of the object, some remaining pixels that belong to the object are still not activated by the CAM. To this end, we propose to further use the concept of entropy to guide the CNN during training. The entropy is a measure of uncertainty, it represents how a CNN is certain or not about its predictions \cite{b22,b26,b21}, the CAM generated by the CNN only highlights the most discriminative features of the object. The pixels that constitute this part have a high prediction probability, i.e., low-entropy. And the other pixels, that are not highlighting by the CAM have a low prediction propability, i.e., high-entropy. By minimizing the entropy of the CAM, the CNN could extend the localization of the object features. 

We can resume our work to the following contributions : 
\begin{enumerate}
\item An entropy guided adversarial model (Dubbed EGA model) that uses both clean examples and adversarial examples to activate more features of the object. 
\item Introduce an entropy loss function to guide the CNN to detect the pixels that belong to the object. 
\item Extensive expriments shown that our EGA model obtained state-of-theart performance in weakly supervised object localization.
\end{enumerate}

\section{Related work}
\label{sec:related work}
\subsection{Weakly Supervised Object Localization} 
Weakly supervised object localization recieved a lot of attention since \cite{b2}, where they used a Global Average Pooling (GAP) layer to generate a Class Activation Map that localizes the object with only label annotation. This map highlights only the most discriminative part of the object, resulting in a tight bounding box. Many works attempt to create new methods to solve this problem. We can devide the proposed solutions into two classes, the first class remove a part or some parts of the image to force the network to detect more features. As in \cite{b1}, where they proposed to use two branches, the first branch generates a map with the discriminative part highlighted, and then removes the pixels of this part, to give it as input to the second branch of the network, the second branch generates another map where other features of the object are highlighted. The two generated maps are then fused by taking the maximum value. Another method that uses a hiding process is \cite{b3}, in which the images are divided into patches, and then for each patch a probability is assigned. At each epoch, some patches are hidden and given as input to the CNN. The CNN hence learns to detect the whole object. During testing, the images are given to the network without removing any patches. \cite{b4} uses either an attention mechanism or a dropout mask to help the network to detect the whole object. \cite{b19} removes the discriminative region and replace it with a patch from another image, hence there is no non-informative pixels in the image and the network could generalize well for both classification and localization. Some other methods do not remove any part of the image and modify the network achitecture to generate different maps from different hierarchies. In \cite{b5}, an attention map generated from high level features map, is used to guide the network to distinguish between the forground and background pixels of the map generated from low level features map. \cite{b11} combines two child classes labels to create a parent class and train the CNN with hierarchical class labels to detect common visual patterns and \cite{b12} generates many CAMs from low and high features map and combine them using a polynomial function. 

For the first approach, when some parts of the object are removed, non discriminative features are activated by the network because they appear too frequently in the dataset. The second approach moreover is not intuitive to apply for complex network architectures as we have to plug some blocks to generate multiple CAMs from different levels of feature maps. In this article, we propose to employ adversarial examples as data augmentation to detect more discriminative features. Minimizing the entropy loss on the CAM will further guide the model during training. 

\subsection{Adversarial learning}
Adversarial learning consists of constructing robust models by training the classifier with adversarial examples \cite{b16}. Some works used it either to improve the robusteness of the model, or to solve other problems; as Madry et al. \cite{b13} where they constructed a robust classifier using a min-max formulation to protect their network against adversarial attacks. Tsipras et al. \cite{b14} relates the benefit of adversarial learning and show that robust classifier learns different features than standard one. \cite{b24} trains a network with both clean examples and adversarial examples to improve the classification accuracy and achieved state of the art accuracy on ImageNet dataset. \cite{b18} employs adversarial learning to improve classification in semi-supervised learning. In \cite{b15}, an adversarial dropout mask is selected using adversarial learning and applied to the network to improve the classification accuracy for both supervised and semi-supervised learning. \cite{b17} proposes a fast training for adversarial training by updating the network parameters and creating an adversarial example in one backward pass. \cite{b25} applies adversarial learning and adversarial dropout to learn discriminative features for unsupervised domain adaptation. In this article, we take advantage of adversarial attacks and use adversarial examples as data augmentation to improve performance of weakly supervised object localization problems.

\subsection{Entropy}
In information theory, entropy is the measure of uncertainty, it tells how a network is uncertain about its prediction. When the prediction of the network is of high porbability, the entropy value is low, and when the prediction is low, the amount of entropy is high. \cite{b26} uses entropy for weakly supervised object detection as an optimization method associated with multiple instance learning to select the object proposals that belong to the object. \cite{b22} shows state of the art performance in Domain Adaptation in Semantic Segmentation using an entropy minimization loss on the target dataset. \cite{b21} considers the entropy as a weighting coefficient to improve weakly supervised learning on Pascal VOC dataset \cite{b41}. Entropy was also used for semi-supervised domain adaptation \cite{b20} and semi-supervised learning \cite{b35}.

\begin{figure*}[t]
\centerline{
 \includegraphics[width=\textwidth]{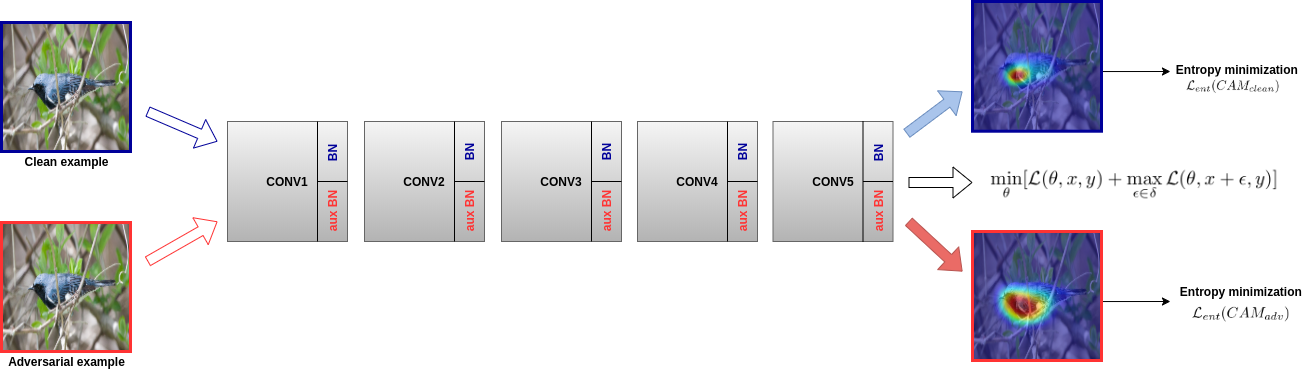}}
\caption{The clean example and adversarial example are fed through the network. The clean example passes through the main batch normalization and the adversarial learning through the auxiliary batch normalization. After the forward pass, we get $CAM_{clean}$ and $CAM_{adv}$ and calculate the shannon entropy loss.}
\label{fig:met}
\end{figure*}  
  
\section{Method}
\label{sec:method} 
In this section, we introduce our Entropy Guided Adversarial model i.e EGD model. In section \ref{sssec:revise}, we review the baseline for WSOL. In \ref{sssec:advlearning}, we briefly present adversarial learning and how to apply it for WSOL problem. In \ref{sssec:ent}, we discuss about the shannon entropy application on CAM, and finally in section \ref{sssec:ent}, we present the loss function used by our network. Our model is illustrated in Fig. \ref{fig:met}.

\subsection{Revising Class Activation Map}
\label{sssec:revise}
Weakly Supervised Object Localization aims at detecting objects without bounding box annotations. One method widely used to localize objects using only label annotations is \cite{b2}. In \cite{b2}, they propose to add to a network, composed of convolutional layers a global average pooling layer before the softmax layer. We denote the last convolutional layer as $f_{k} (h,w)$. The GAP layer is applied to the last convolutional layer and output a vector where each unit is the average of each feature map :
\begin{eqnarray}
  F_{k} =  \sum_{(h,w)}f_{k} (h,w) 
  \label{eq:6}
\end{eqnarray}
The weights of the generated vector are then multiplied by the features map of the last convolutional layer and are given as input to the softmax layer as: 
\begin{eqnarray}
  S_{c} =  \sum_{k}a_{k}^{c}\sum_{(h,w)}f_{k} (h,w) 
  \label{eq:7}
\end{eqnarray}
where $a_{k}^{c}$ is the weight of the corresponding class c which indicates the importance of $F_{k}$ for class c.

To generate a map that indicates the importance of each pixel for the corresponding class, we summed up the weighted features map as:
\begin{eqnarray}
  CAM_{c}(h,w) = \sum_{k}a_{k}^{c}f_{k} (h,w)
  \label{eq:8}
\end{eqnarray}
where $CAM_{c}(h,w)$ is the class activation map generated for class c. The CAM activates the features that discriminate the most the object.

\begin{figure*}[t]
  \centerline{\includegraphics[width=\textwidth]{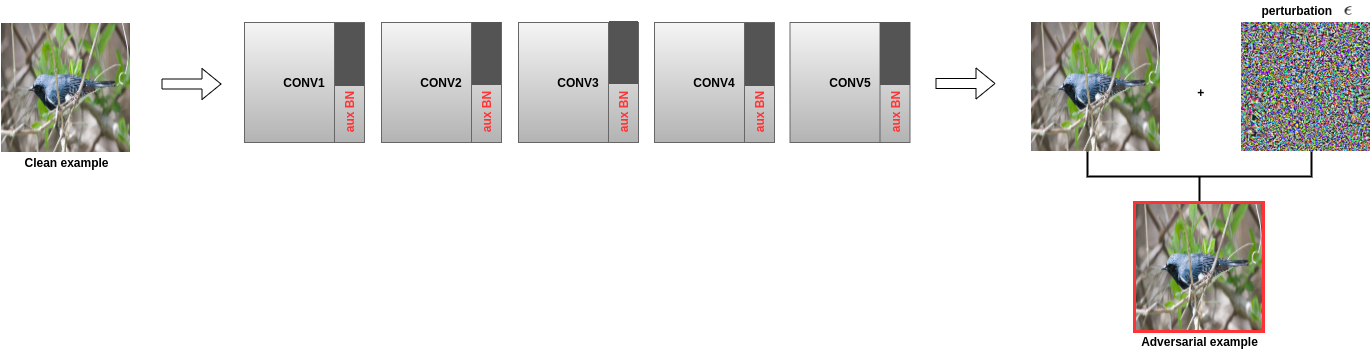}}
\caption{The clean example is given as input to the network to create a perturbation $\epsilon$ that maximizes the loss function. We drop the main batch norm and use the auxiliary batch norm to generate the adversarial example. }
\label{fig:adv} 

\end{figure*} 

\subsection{Adversarial learning}
\label{sssec:advlearning}
To train a CNN, we use stocastic gradient descent method to minimize an objective function according to the parameters $\theta$ of the network:  
\begin{eqnarray} 
  \min_{\theta}\mathcal{L} (\theta, x, y) 
  \label{eq:1}
\end{eqnarray}
where $x \in \mathcal{X}$ is an input sample associated with its ground truth label $ y \in \mathcal{Y}$ and $\mathcal{L}(.,.,.)$ is the objective function i.e loss function. 
\begin{figure}[t]
\centerline{
  \includegraphics[width=0.5\textwidth]{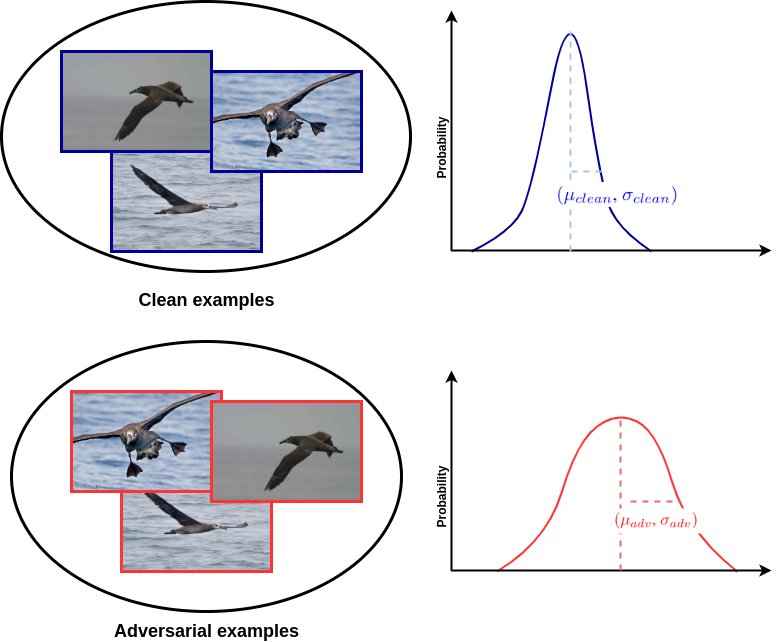}}

\caption{The difference between the distribution of clean examples and adversarial examples.}
  
\label{fig:dist}
\end{figure}

To construct a model that is robust to adversarial attacks, we train the classifier with adversarial examples by adding a small perturbation to the image to fool the network \cite{b16,b13}. The resulted sample should be close to the original image, and maximize the loss function instead of minimizing it as we usually do for standard training, the objective function hence becomes : 
\begin{eqnarray}
  \min_{\theta} [ \max_{ \substack{\epsilon \in \delta} } \mathcal{L} (\theta, x+\epsilon, y) ]
  \label{eq:2}
\end{eqnarray}
where $\epsilon$ is the adversarial perturbation i.e noise and $\delta$ is the allowable set of perturbations which ensure that adversarial image is close to the clean one.

Goodfellow et al \cite{b16} advice the use of a mixture of clean examples and adversarial examples to train the model. Adversarial examples act as data augmentation and associated with clean examples regularize the network. As suggested in Goodfellow et al work the objective function is: 
\begin{eqnarray}
  \min_{\theta} [\mathcal{L} (\theta, x, y) + \max_{\substack{\epsilon \in \delta}} \mathcal{L} (\theta, x+\epsilon, y) ]
  \label{eq:3}
\end{eqnarray}
Clean samples and adversarial samples can be considered as two different datasets, because they have two different disctributions, hence the network could not generalize well on clean data during the inference. This is due to the distribution mismatch between the two type of datasets \cite{b24}. As shown in Fig. \ref{fig:dist}, the mean and the standard deviation of the ditribution of clean data is different from the one of adversarial data. 

To solve this problem, we follow the work of \cite{b24} and add an auxiliary batch normalization to the network. Batch normalization (BN) is a technique used to normalize the input data \cite{b36}. As we have two datasets distributed differently, we use one batch norm for each type of dataset i.e we feed each sample to its corresponding batch norm. Specifically, for each clean mini batch $x_{clean}$, we generate its corresponding adversarial mini-batch $x_{adv}$ using the auxiliary batch norm layer. The process of generating the adversarial example is shown in Fig. \ref{fig:adv}. We feed a clean sample to the network to calculate the perturbation that maximizes the loss function using Eq (\ref{eq:2}). The network parameters are not updating during this process. We then pass the clean mini batch and adversarial mini batch through the network to compute the loss, the clean mini batch pass through the main batch normalization and the adversarial mini batch pass through the auxiliary batch normalization. After the forward pass, the model generates two maps; $CAM_{clean}$ and $CAM_{adv}$ for clean sample and adversarial sample respectively. After that, we calculate the loss function and update the network parameter. During inference, we test our model only on clean data, the auxialiary batch norm is dropped and only the main batch norm is used.

\subsection{Entropy minimization}
\label{sssec:ent}
Unlike previous wroks \cite{b1,b2}, where they generate the CAM during the test stage, we generate our CAMs after each forward pass to guide the CNN for localization during the training. Our adversarial model using the main and auxiliary batch norm layer generates two different CAMs; $CAM_{clean}$ and $CAM_{adv}$ for the clean example and adversarial example respectively. These CAMs activate different parts of the object but still do not activate the whole object. We propose to use the shannon entropy to remedy to this problem.

The Shannon entropy measure the amount of uncertainty \cite{b22}, when it is applied to the CAM, a pixel with a low entropy means a high prediction probability and a pixel with a high entropy means a low prediction probability. The maps generated by the network highlight the most discriminative part i.e these pixels have a low entropy; the other pixels that are not highlighted by the maps but belongging to the object have a high entropy. By minimizing the entropy of the CAM, we force the CNN to activate the pixels that belong to the object but not highlighted by the map

Given the generated CAMs, $CAM_{clean}$ and $CAM_{adv}$, we calculate the entropy loss for one $CAM$ as :
\begin{eqnarray}
  \mathcal{L}_{ent}(CAM) = -\sum_{(h,w)} P_{CAM^{(h,w)}}\log{P_{CAM^{(h,w)}}}
  \label{eq:4}
\end{eqnarray}
where $P_{CAM^{(h,w)}}$ represent the probability of the pixel $(h,w)$.  
$\mathcal{L}_{ent}(CAM)$ is the sum of all entropies pixels i.e we maximize the predictions on the pixels that are not activated by the CAM.

\subsection{Loss function}
\label{sssec:loss}
After feeding the network with clean and adversarial sample, we generate $CAM_{clean}$ and $CAM_{adv}$, we then calculate the loss entropy from the two maps. We jointly optimize the adversarial learning loss function and entropy loss function to optimize the network parameters by summing Eq (\ref{eq:3}) and Eq (\ref{eq:4}). The final loss function is defined as below :
\begin{eqnarray}
  \min_{\theta} [\mathcal{L} (\theta, x, y) + \max_{\substack{\epsilon \in \delta}} \mathcal{L} (\theta, x+\epsilon, y) ] + \lambda_{CAM_{clean}}\mathcal{L}_{ent}(CAM_{clean}) + \lambda_{CAM_{adv}}\mathcal{L}_{ent}(CAM_{adv}) 
  \label{eq:5}
\end{eqnarray}
where $\lambda_{CAM_{clean}}$ and $\lambda_{CAM_{adv}}$ is the weighting factor controlling the importance of $\mathcal{L}_{ent}(CAM_{clean})$ and $ \mathcal{L}_{ent}(CAM_{adv})$, respectively.

Because of the perturbation on the adversarial example, $CAM_{adv}$ has more activated pixels than $CAM_{clean}$. To this end, we set $\lambda_{CAM_{clean}} > \lambda_{CAM_{adv}}$, this setting pushes the classifier to be more severe with clean examples by forcing the $CAM_{clean}$ to activate more pixels.

\section{Experiments}
\label{sec:experiments}
\subsection{Experimental Setup}
\paragraph{Datasets} We evaluate our method on two commonly used datasets for WSOL i.e CUB-200-2011 \cite{b6} and ILSVRC \cite{b9,b10}. We further evaluation EGA model on OpenImages, a fresh new dataset proposed by \cite{b23} for WSOL. CUB-200- 235 2011 consists of 200 bird species, this dataset contains 11,788 images with 5,994 images for training and 5,794 for testing. ILSVRC 2016, is a large scale dataset of 1000 classes, that comprise 1.2 million for training, and 5,000 images for the validation set that we use for testing. OpenImages has 100 classes, it contains 29 819 for training, 2 500 for validation, and 5 000 images for testing.

\paragraph{Evaluation metrics} For classification evaluation, we use the Top-1 classification accuracy, which indicates that a prediction is correct when the prediction of the model is equal to the ground-truth class. For localization, as CUB and ILSVRC datasets have bounding boxes annotations, we use three evaluation metrics : Top-1 localization accuracy \cite{b9}, Correct Localization \cite{b38} (CorLoc) rate and MaxBoxAccV2 \cite{b23}. Top-1 localization accuracy counts a localization as correct when the predicted class is correct and the predicted bounding box has an Intersaction Over Union (overlap) with the ground truth bounding box greater than 0.5. Correct Localization (CorLoc) is the localization performance whether or not the predicted class is correct. MaxBoxAccV2 is a new metric proposed by \cite{b23}, which is an improved version of CorLoc; we average the results of the IOU between the predicted box and the ground truth box accross 0.3, 0.5, 0.7. OpenImages provides pixel-wise annotations, hence we use pixel average precision (PxAP) metric \cite{b23} for localization, which measure the pixelwise precision and recall trade-off.

\paragraph{Experimental details} We use for training VGGnet \cite{b8} and GoogLeNet \cite{b40}, We follow the same setting as \cite{b2}, we remove the layers after conv5-3 (from pool5 to prob) of the VGG-16 network and the last inception block of GoogLeNet. We then add two convolutional layers with kernel size 3 $\times$  3, stride 1, pad 1 with 1024 units, and a convolutional layer of size 1 $\times$ 1, stride 1 with 1000 units for ILSVRC, 200 units for CUB-200-2011 and 100 units for OpenImages. For training, input images are resized to 256 $\times$  256, then randomly cropped to 224 $\times$ 224. Both backbone networks are fine-tuned on the pre-trained weights of ILSVRC. During inference, we resized images to 224 $\times$ 224 to find the whole objects and for classification for CUB and ILSVRC datasets, we average the scores from the softmax layer with 10 crops.

\begin{table}[t!]

\centering
 \begin{tabular}{|c|c|c|} 
 \hline
 Method & top1 cls-err & top1 loc-err \\
 \hline
 GoogLeNet-CAM \cite{b2} & 26.2 & 58.94 \\
 GoogLeNet-SPG \cite{b5} & - & 53.36 \\
 GoogLeNet-ADL \cite{b4} & \textbf{25.45} & \textbf{46.94} \\
 GoogLeNet-DANet \cite{b11} & 28.8 & 50.55 \\
 \textbf{GoogLeNet-EGA (ours)} & 27.89 & 54.26 \\
 \hline
 VGGnet-CAM \cite{b2} & 23.4 & 55.85 \\
 VGGnet-ACoL \cite{b1} & 28.1 & 54.08 \\
 VGGnet-SPG \cite{b5} & 24.5 & 51.07 \\
 VGGnet-CutMix \cite{b19} & - & 47.47 \\
 VGGnet-ADL \cite{b4} & 34.73 & 47.64 \\
 VGGnet-DANet \cite{b11} & 24.6 & 47.48 \\
 VGGnet-CCAM \cite{b12} & 26.8 & 49.93 \\ 
 \textbf{VGGnet-EGA (ours)} & \textbf{21.87} & \textbf{40.84} \\
 \hline
 \end{tabular}
 \caption{Comparison to the state-of-the-art performance on the CUB dataset.}
 \label{tab:1}
\end{table}
\begin{table}[t!]

\centering
 \begin{tabular}{|c|c|c|} 
 \hline
 Method & top1 cls-err & top1 loc-err \\
 \hline
 GoogLeNet-Backprop \cite{37} & - & 61.31 \\
 GoogLeNet-CAM \cite{b2} & 35.0 & 56.40 \\
 GoogLeNet-HaS \cite{b3} & - & 54.53 \\
 GoogLeNet-ACoL \cite{b1} & 29.0 & 53.28 \\ 
 GoogLeNet-SPG \cite{b5} & - & 51.40 \\
 GoogLeNet-ADL \cite{b4} & \textbf{27.17} & 51.29 \\ 
 GoogLeNet-DANet \cite{b11} & 27.5 & 52.47 \\
 \textbf{GoogLeNet-EGA (ours)} & 27.42 & \textbf{50.17} \\
 \hline
 VGGnet-Backprop \cite{b37} & - & 61.12 \\
 VGGnet-CAM \cite{b2} & 33.4 & 57.20 \\
 VGGnet-ACoL \cite{b1} & 32.5 & 54.17 \\
 VGGnet-CutMix \cite{b19} & - & 56.45 \\
 VGGnet-ADL \cite{b4} & 30.52 & 55.08 \\
 NL-CCAM \cite{b12} & \textbf{27.7} & \textbf{49.83} \\ 
 \textbf{VGGnet-EGA (ours)} & 29.36 & 52.69 \\
 \hline
 \end{tabular}
 \caption{Comparison to the state-of-the-art performance on the ILSVRC validation set.}
 \label{tab:2}
\end{table}

\subsection{Comparison with the state-of-the-arts} 
We compare our EGA model to the state of the arts on CUB, ILSVRC and the new proposed dataset OpenImages. The results are shown in Tab. \ref{tab:1}, Tab. \ref{tab:2} and Tab. \ref{tab:3}, respectively.

\paragraph{CUB} As shown in Tab. \ref{tab:1}, with VGG model, our model outperform by far all the previous state of the arts in both classification and localization with 21.87\% for classification and 40.84\% for localization. We improved our baseline VGGnet-CAM by 1.53\% and 15.01\% for classification and localization respectively. We also surpass VGGnet-CCAM, the current state of the art for localization with a large margin of 9.09\% for localization and margin of 4.93\% for classification. 

With a GoogLeNet backnone, our EGA model achieved 27.89\% and 57.26\% for classification and localization respectively. We did not achieve a new state of the art with GoogLeNet architecture but we surpass our baseline GoogLeNetCAM for localization with 4.68\%. We argue that GoogLeNet compared to VGG backbone is deeper and hence, needs a larger dataset as ILSVRC dataset to achieve good performance, as we add some perturbation to the samples, the network needs more data to improve its classification and localization accuracy.
 
\paragraph{ILSVRC} In Tab. \ref{tab:2}, with VGG backbone, we achieved 29.36\% and 52.69\% for classification and localization respectively on ILSVRC dataset, we surpass the basline with a difference of 4.04\% and 4.51\%, we also suprpass all the state of the art method for both classification and localization, except for NL-CCAM which outperforms our method with a margin of 1.66\% and 2.86\% for classification and localization respectively.

With GoogLeNet architecture, EGA model achieved 27.42\% and 50.17\% for classification and localization respectivaly.
As supposed earlier, our method with GoogLeNet backbone with a larger dataset outperform all state of the art for localization, we improved our basline with a margin of 7.58\% and 6.23\% for classification and localization respectively. We also surpass the current state of the art ADL with a difference of 1.12\% for localization with a good classification accuracy. 

\begin{figure*}[t]

  \centerline{
  \includegraphics[width=\textwidth]{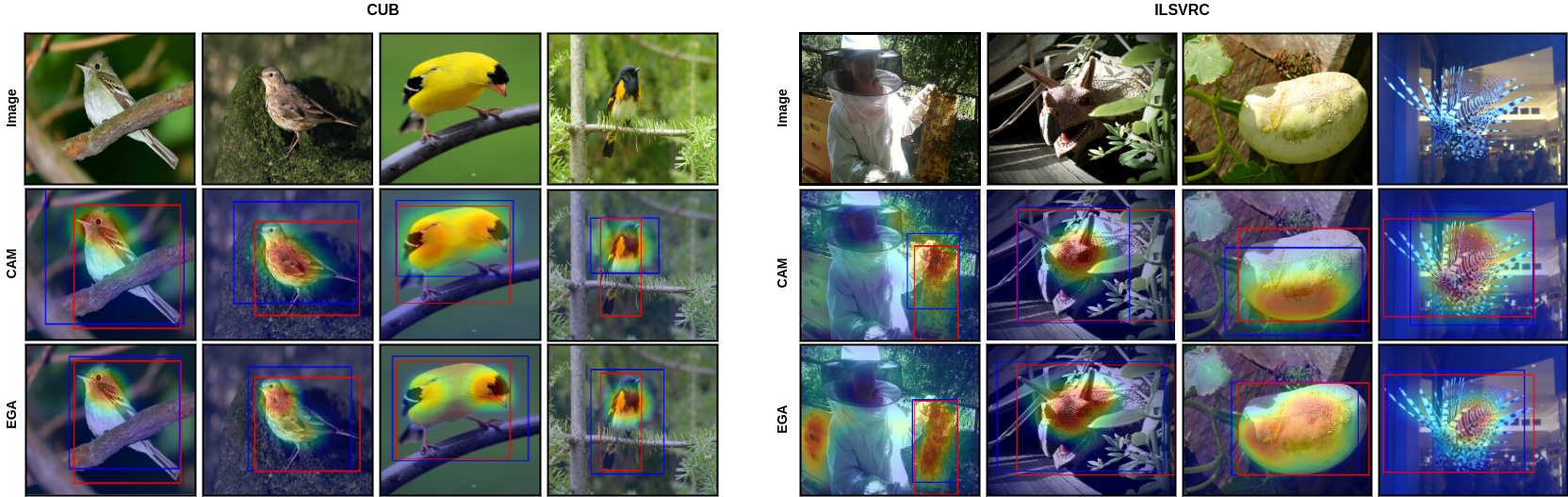}}

\caption{Comparison of our EGA method to CAM method. Our method activates more object’s features than CAM and generates tighter bounding boxes. Ground-truth bounding boxes are in red and the predicted are in blue.}
  
\label{fig:cams}
\end{figure*}

In Fig. \ref{fig:cams}, we compare the bounding boxes generated by CAM method \cite{b2} and bounding boxes generated by our EGA model. Our method activates more object’s features than the baseline \cite{b2}.

\begin{table}[t!]
\centering
 \begin{tabular}{|c|c|c|} 
 \hline
 Method & top1 cls-err & top1 loc-err \\
 \hline
 GoogLeNet-CAM \cite{b2} & 63.4 & \textbf{36.8} \\
 GoogLeNet-HaS \cite{b3} & \textbf{31.6} & 41.9 \\
 GoogLeNet-ACoL \cite{b1} & 59.3 & 42.8 \\ 
 GoogLeNet-SPG \cite{b5} & 53.4 & 37.7 \\
 GoogLeNet-ADL \cite{b4} & 53.4 & 43.2  \\ 
 GoogLeNet-CutMix \cite{b19} & 46.9 & 37.5 \\
 \textbf{GoogLeNet-EGA (ours)} & 33.4 & 37.35 \\
 \hline
 VGGnet-CAM \cite{b2} & 32.7 & 41.7 \\
 VGGnet-HaS \cite{b3} & 40.0 & 41.9 \\
 VGGnet-ACoL \cite{b1} & 31.8 & 45.7 \\
 VGGnet-SPG \cite{b5} & \textbf{28.3} & 41.7 \\
 VGGnet-ADL \cite{b4} & 33.9 & 41.3 \\
 VGGnet-CutMix \cite{b19} & 31.9 & 41.9 \\
 \textbf{VGGnet-EGA (ours)} & 30.0 & \textbf{38.21} \\
 \hline
 \end{tabular}
 \caption{Comparison to the state-of-the-art performance on OpenImages dataset with PxAP metric.}
 \label{tab:3}
\end{table}

\begin{figure}[t]

  \centerline{
  \includegraphics[width=0.5\textwidth]{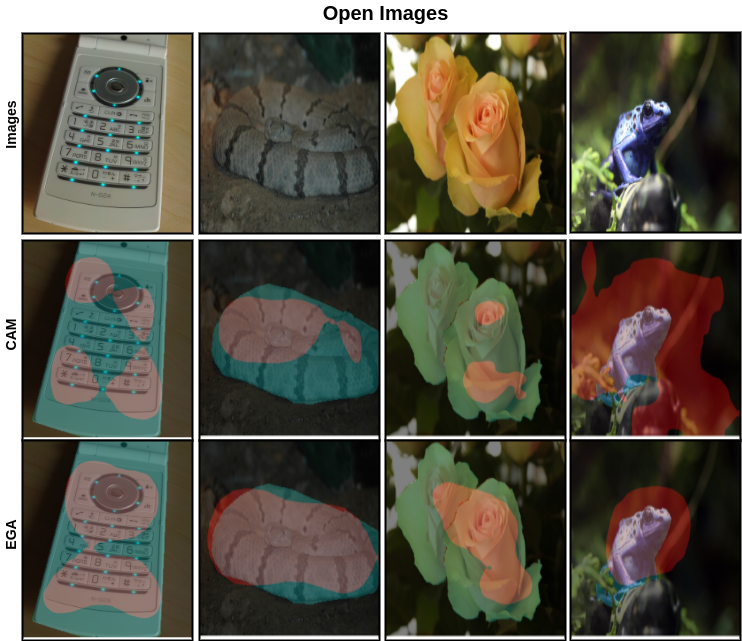}}

\caption{Comparison of our EGA method to CAM method on OpenImages dataset. The ground truth mask is in blue and the predicted mask is in red.}
  
\label{fig:openimages}
\end{figure}

\paragraph{OpenImages} As OpenImages is a new dataset proposed by \cite{b23}, we compared our method to the results reported in \cite{b23}. In Tab. \ref{tab:3}, with VGG backbone, our method achevies 30.0\% for classification and 38.21\% for localization, we still improve the baseline CAM with 2.7\% for classification and 3.49\% for localization. We further surpass ADL the current state of the art on OpenImages dataset with a margin of 3.9\% for classification and a margin of 3.09\% for localization. Hence we achieved a new state of the localization on OpenImages dataset with a small drop for classification. In Fig. \ref{fig:openimages}, our EGA model detects more foreground features and less background features compared to VGGnet-CAM \cite{b2}.

With GooleNet, our method achieves 33.4\% for classification and 37.35\% for localization, we have a slight decrease compared to the baseline CAM of 0.55\% for localization, but we outperform HaS, ACoL, SPG, ADL and CutMix with a margin of 4.55\%, 5.45\%, 0.35\%, 5.85\%, and 0.15\%, respectively.

\paragraph{MaxBoxAccV2} we further evaluate our method with the new evaluation metric MaxBoxAccV2 proposed by \cite{b23}. The results are shown in Tab. \ref{tab:4}. For CUB dataset with a VGG backbone, we still surpass all the sate of the art CAM, HaS, ACoL, SPG and CutMix with a difference of 0.23\%, 0.53\%, 6.53\%, 7.63\% and 1.63\%, except for ADL where there is a difference of 2.37\%.
 For ILSVRC dataset, we achieved a new state of the art for localization by surpassing all previous works with a localization accuracy of 38.22\%. 

\paragraph{CorLoc} We also evaluate our method with Correct Localization metric, a highly used evaluation metric in WSOL \cite{b38}. As shown in Tab. \ref{tab:5}, EGA method improves CAM, HaS, ACoL and SPG methods by 6.17\%, 3.63\%, 1.87\% and 0.14\%. We also outperform the other state of the arts methods, except for NL-CCAM where we have a slight increase of 0.4\%. NL-CCAM applied a Non-Local module to the VGG backbone and hence, this method changes the architecture of the VGG and does not use the original VGG backbone. Our method does not change the backbone of VGG or GoogleNet and we still have competetive results with NL-CCAM. When we compare CCAM method applied to the original basline i.e VGGnet-CCAM we outperform it by 1.25\%.

\begin{table}[t!]

\centering
 \begin{tabular}{|c|c|c|} 
 \hline
 Method & ILSVRC & CUB \\
 \hline
 VGGnet-CAM \cite{b2} & 40.0 & 36.3 \\
 VGGnet-HaS \cite{b3} & 39.4 & 36.6 \\
 VGGnet-ACoL \cite{b1} & 42.6 & 42.6 \\
 VGGnet-SPG \cite{b5} & 40.1 & 43.7 \\
 VGGnet-ADL \cite{b4} & 40.2 & \textbf{33.7} \\
 VGGnet-CutMix \cite{b19} & 40.6 & 37.7 \\
 \textbf{VGGnet-EGA (ours)} & \textbf{38.22} & 36.07 \\
 \hline
 \end{tabular}
 \caption{Comparison to the state-of-the-art performance on the ILSVRC, CUB datasets with MaxBoxAccV2 metric.}
 \label{tab:4}
\end{table}

\begin{table}[t!]

\centering
 \begin{tabular}{|c|c|} 
 \hline
 Method & ILSVRC  \\
\hline
 AlexNet-GAP \cite{b2} & 45.01 \\
 AlexNet-HaS \cite{b3} & 41.25 \\
 AlexNet-GAP-ensemble \cite{b2} & 42.98 \\
 AlexNet-HaS-ensemble \cite{b3} & 39.67 \\
 GoogLeNet-GAP \cite{b2} & 41.34 \\
 GoogLeNet-HaS \cite{b3} & 38.8 \\
 GoogLeNet-ACoL \cite{b1} & 37.04 \\
 GoogLeNet-SPG \cite{b5} & 35.31 \\
 VGGnet-CCAM \cite{b12} & 36.42 \\
 NL-CCAM \cite{b12} & \textbf{34.77} \\
 \textbf{GoogLeNet-EGA (ours)} & 35.17 \\
 \hline
 \end{tabular}
 \caption{Comparison to the state-of-the-art performance on the ILSVRC validation set with CorLoc metric.}
 \label{tab:5}
\end{table}

\subsection{Ablation Study}
\begin{figure*}[t]

  \includegraphics[width=\textwidth]{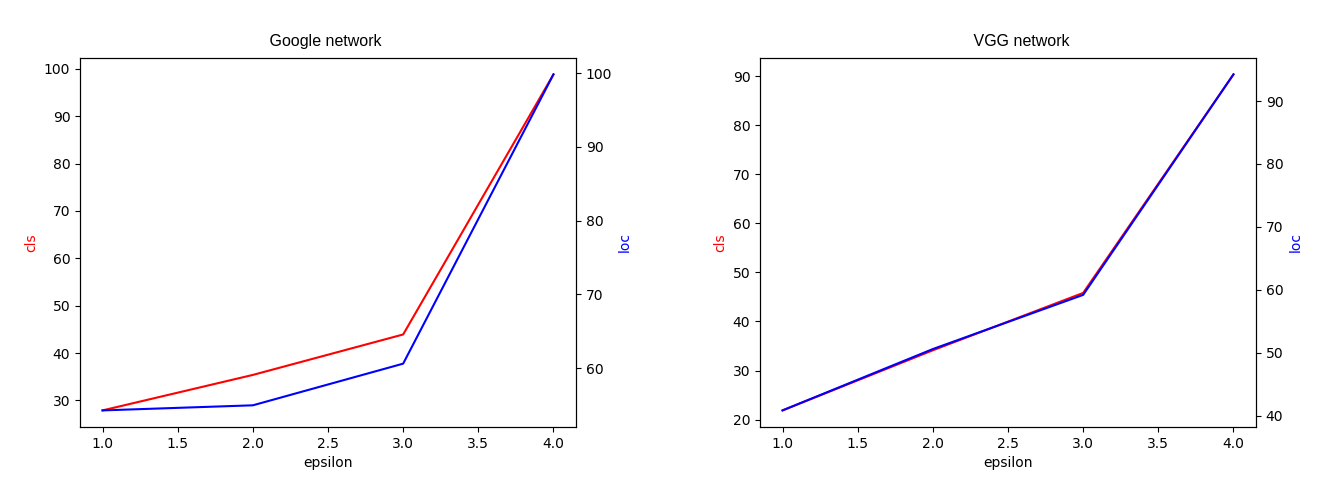}

\caption{The effect of different perturbation strength on the classification and localization accuracy.}
  
\label{fig:epsilon}
\end{figure*}

In this section, we perform ablation study on CUB dataset with VGGnet and GoogLeNet networks. We firstly evaluate the effect of each contribution over the baseline, then the effect of different perturbation values on localization and classification, and finally, how $\lambda_{CAM_{clean}}$ and $\lambda_{CAM_{adv}}$ influence the results of our method. The results are reportes in Tab. \ref{tab:6} and Tab. \ref{tab:7} and Fig. \ref{fig:epsilon}.

\paragraph{Effect of AL and Entropy} As shown in Tab. \ref{tab:6}, with VGGnet backbone, using clean examples and adversarial examples improves greatly the baseline i.e 1.55\% for classifcation and 14.92\% for localization. When we further apply the entropy, we improve the localization accuracy by 0.09\% with a little drop in classification of 0.02\%. With GoogLeNet backbone, using adversarial learning improves the baseline with a margin of 4.12\% for localization, however it drops the classification accuracy by 1.73\%. When the entropy is further applied we improve both the classification and the localization by 0.04\% and 0.56\%, respectively  

\paragraph{Adversarial Attacker Strength} We show the effect of Projected Gradient Descent (PGD) \cite{b13} attackers with different perturbation values. we train both VGG network and GoogLeNet network on CUB dataset and as \cite{b24}, we use perturbations $\epsilon$ ranging from 1 to 4 with an iteration of $n = \epsilon + 1$, except when $\epsilon = 1 $ we set the number of iteration to 1. As shwon in Fig. \ref{fig:epsilon}, the bigger the perturbation, the higher is the error for both classification and localization. We get the best results with $\epsilon = 1$ and $n = 1$. This is obvious as our goal is not to build a robust model, but using adversarial learning as a way to activate more relevant features in the image.

\paragraph{Regularization factors} We further show the effect of the regularization factors $\lambda_{CAM_{clean}}$ and $\lambda_{CAM_{adv}}$ on the method, As shown in Tab. \ref{tab:7}, with VGGnet backbone, we got the best results with $\lambda_{CAM_{clean}} = 1$ and $\lambda_{CAM_{adv}} = 0.01$. By selecting the right values for $\lambda_{CAM_{clean}}$ and $\lambda_{CAM_{adv}}$, entropy improves the localization accuracy.

\begin{table}[t!]

\centering
 \begin{tabular}{|c|c|c|} 
 \hline
 Method & top1 cls-err & top1 loc-err  \\
\hline
 VGGnet-CAM \cite{b2} & 23.4 & 55.85 \\
 VGG + Adversarial learning & \textbf{21.85} & 40.93 \\
 VGG + Adversarial learning + Entropy & 21.87 & \textbf{40.84} \\
 \hline
 GoogLeNet-CAM \cite{b2} & 26.2 & 58.94 \\
 GoogLeNet + Adversarial learning & 27.93 & 54.82 \\
 GoogLeNet + Adversarial learning + Entropy & \textbf{27.89} & \textbf{54.26} \\
 \hline
 \end{tabular}
 \caption{The effect of adding adversarial training and entropy loss to the baseline CAM.}
 \label{tab:6}
\end{table}

\begin{table}[t!]

\centering
 \begin{tabular}{|c|c|c|} 
 \hline
 Method & top1 cls-err & top1 loc-err \\
 \hline
 $\lambda_{CAM_{clean}} = 3$, $\lambda_{CAM_{adv}} = 1$ & 22.59 & 41.94 \\
 $\lambda_{CAM_{clean}} = 1$, $\lambda_{CAM_{adv}} = 0.01$ & \textbf{21.87} & \textbf{40.84} \\
 $\lambda_{CAM_{clean}} = 0.1$, $\lambda_{CAM_{adv}} = 0.01$ & 22.14 & 41.54 \\ 
 $\lambda_{CAM_{clean}} = 0.01$, $\lambda_{CAM_{adv}} = 0.002$ & 21.88 & 41.23 \\
 $\lambda_{CAM_{clean}} = 0.001$, $\lambda_{CAM_{adv}} = 0.0002$ & 22.11 & 41.09  \\ 
 \hline
 \end{tabular}
 \caption{the effect of $\lambda_{CAM_{clean}}$ and $\lambda_{CAM_{adv}}$ on the results.}
 \label{tab:7}
\end{table}

\section{Conclusion}
In this paper, we proposed to take advantage of adversarial learning and entropy to improve WSOL performance. To do this, we train the model with clean examples and adversarial examples. By introducing some perturbations to the images, adversarial examples act as data augmentation and regularize the network, resulting in the activation of more relevant features. Furthermore,  applying entropy minimization on the CAMs generated by the network, guides it during the training by forcing the pixels considered not relevant by the model to have a low entropy, and hence a higher prediction. Extensive experiments demonstrate that our EGA model obtained state of the arts on the three most used benchmark CUB, ILSVRC and OpenImages datasets.

\section*{References}

\bibliography{mybibfile}

\end{document}